\documentclass{article}

\usepackage[letterpaper,margin=1in]{geometry}

\usepackage[utf8]{inputenc} 
\usepackage[T1]{fontenc}    
\usepackage{hyperref}       
\usepackage{url}            
\usepackage{booktabs}       
\usepackage{amsfonts}       
\usepackage{nicefrac}       
\usepackage{microtype}      

\usepackage{amsmath, amssymb, amsthm, bbm}
\usepackage{color}
\usepackage{comment}
\usepackage{multirow}
\usepackage{graphicx}
\usepackage{subcaption}
\usepackage{listings}
\usepackage{paralist}
\usepackage{float}
\usepackage{caption}

\newcommand*\samethanks[1][\value{footnote}]{\footnotemark[#1]}

\newcommand{\imgpath}[0]{figs/}
\newcommand{\appendixorsupp}[1]{Appendix~\ref{#1}}

\usepackage{tikz}
\usetikzlibrary{shapes,backgrounds,arrows,decorations.pathreplacing}

\title{Learning to Compose Domain-Specific Transformations\\for Data Augmentation}

\author{
    Alexander J. Ratner\thanks{Authors contributed equally},~~Henry R. Ehrenberg\samethanks,~~Zeshan Hussain,\\
    Jared Dunnmon,~~Christopher R\'{e}\\
    Stanford University\\
    \texttt{\{ajratner,henryre,zeshanmh,jdunnmon,chrismre\}@cs.stanford.edu}\\
}

\begin{document}

\maketitle

\begin{abstract}
  Data augmentation is a ubiquitous technique for increasing the size of labeled training sets by leveraging task-specific data transformations that preserve class labels.
While it is often easy for domain experts to specify individual transformations, constructing and tuning the more sophisticated compositions typically needed to achieve state-of-the-art results is a time-consuming manual task in practice.
We propose a method for automating this process by learning a generative sequence model over user-specified transformation functions using a generative adversarial approach.
Our method can make use of arbitrary, non-deterministic transformation functions, is robust to misspecified user input, and is trained on unlabeled data.
The learned transformation model can then be used to perform data augmentation for any end discriminative model.
In our experiments, we show the efficacy of our approach on both image and text datasets, achieving improvements of 4.0 accuracy points on CIFAR-10, 1.4 F1 points on the ACE relation extraction task, and 3.4 accuracy points when using domain-specific transformation operations on a medical imaging dataset as compared to standard heuristic augmentation approaches.
\end{abstract}

\section{Introduction}

Modern machine learning models, such as deep neural networks, may have billions of free parameters and accordingly require massive labeled data sets for training.
In most settings, labeled data is not available in sufficient quantities to avoid overfitting to the training set.
The technique of artificially expanding labeled training sets by transforming data points in ways which preserve class labels -- known as \textit{data augmentation} -- has quickly become a critical and effective tool for combatting this labeled data scarcity problem.
Data augmentation can be seen as a form of \textit{weak supervision}, providing a way for practitioners to leverage their knowledge of invariances in a task or domain.
And indeed, data augmentation is cited as essential to nearly every state-of-the-art result in image classification~\cite{ciresan80deep,dosovitskiy2015discriminative,graham2014fractional,SajjadiJT16a} (see \appendixorsupp{sec:lit-review}), and is becoming increasingly common in other modalities as well~\cite{lu2006enhancing}.

Even on well studied benchmark tasks, however, the choice of data augmentation strategy is known to cause large variances in end performance and be difficult to select~\cite{graham2014fractional,dosovitskiy2015discriminative}, with papers often reporting their heuristically found parameter ranges~\cite{ciresan80deep}.
In practice, it is often simple to formulate a large set of primitive transformation operations, but time-consuming and difficult to find the parameterizations and compositions of them needed for state-of-the-art results.
In particular, many transformation operations will have vastly different effects based on parameterization, the set of other transformations they are applied with, and even their particular order of composition.
For example, brightness and saturation enhancements might be destructive when applied together, but produce realistic images when paired with geometric transformations.

Given the difficulty of searching over this configuration space, the de facto norm in practice consists of applying one or more transformations in random order and with random parameterizations selected from hand-tuned ranges.
Recent lines of work attempt to automate data augmentation entirely, but either rely on large quantities of labeled data~\cite{baluja2017adversarial,mirza2014conditional}, restricted sets of simple transformations~\cite{fawzi2016adaptive,hauberg2016dreaming}, or consider only local perturbations that are not informed by domain knowledge~\cite{baluja2017adversarial,miyato2015distributional} (see Section~\ref{sec:related-work}).
In contrast, our aim is to directly and flexibly leverage domain experts' knowledge of invariances as a valuable form of weak supervision in real-world settings where labeled training data is limited.

In this paper, we present a new method for data augmentation that directly leverages user domain knowledge in the form of transformation operations, and automates the difficult process of composing and parameterizing them.
We formulate the problem as one of learning a generative sequence model over black-box \textit{transformation functions (TFs)}: user-specified operators representing incremental transformations to data points that need not be differentiable nor deterministic.
For example, TFs could rotate an image by a small degree, swap a word in a sentence, or translate a segmented structure in an image (Fig. \ref{fig:tfs}). We then design a generative adversarial objective~\cite{goodfellow2014generative} which allows us to train the sequence model to produce transformed data points which are still within the data distribution of interest, using unlabeled data.
Because the TFs can be stochastic or non-differentiable, we present a reinforcement learning-based training strategy for this model.
The learned model can then be used to perform data augmentation on labeled training data for any end discriminative model.

Given the flexibility of our representation of the data augmentation process, we can apply our approach in many different domains, and on different modalities including both text and images.
On a real-world mammography image task, we achieve a 3.4 accuracy point boost above randomly composed augmentation by learning to appropriately combine standard image TFs with domain-specific TFs derived in collaboration with radiology experts.
Using novel language model-based TFs, we see a 1.4 F1 boost over heuristic augmentation on a text relation extraction task from the ACE corpus.
And on a 10\%-subsample of the CIFAR-10 dataset, we achieve a 4.0 accuracy point gain over a standard heuristic augmentation approach and are competitive with comparable semi-supervised approaches.
Additionally, we show empirical results suggesting that the proposed approach is robust to misspecified  TFs.
Our hope is that the proposed method will be of practical value to practitioners and of interest to researchers, so we have open-sourced the code at \url{https://github.com/HazyResearch/tanda}.

\section{Modeling Setup and Motivation}
\label{sec:setup}

\begin{figure}
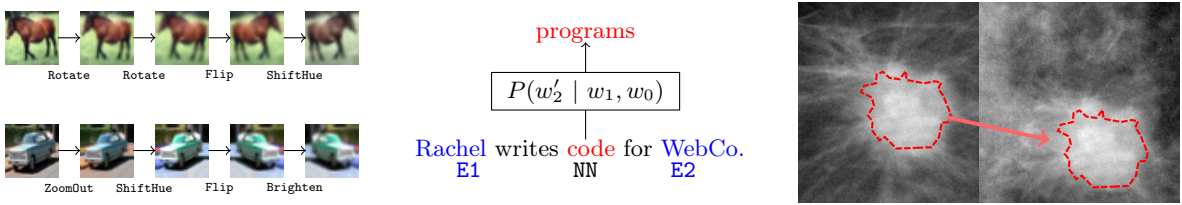

\centering
\include{tanda_tfs_fig}
\caption{Three examples of transformation functions (TFs) in different domains: Two example sequences of incremental image TFs applied to CIFAR-10 images (\textit{left}); a conditional word-swap TF using an externally trained language model and specifically targeting nouns (\texttt{NN}) between entity mentions (\texttt{E1,E2}) for a relation extraction task (\textit{middle}); and an unsupervised segementation-based translation TF applied to mass-containing mammography images (\textit{right}).}
\label{fig:tfs}
\end{figure}


In the standard data augmentation setting, our aim is to expand a labeled training set by leveraging knowledge of class-preserving transformations.
For a practitioner with domain expertise, providing individual transformations is straightforward.
However, high performance augmentation techniques use \textit{compositions} of finely tuned transformations to achieve state-of-the-art results~\cite{dosovitskiy2015discriminative,ciresan80deep,graham2014fractional}, and heuristically searching over this space of all possible compositions and parameterizations for a new task is often infeasible.
Our goal is to automate this task by learning to compose and parameterize a set of user-specified transformation operators in ways that are diverse but still preserve class labels.

In our method, transformations are modeled as sequences of incremental user-specified operations, called transformation functions (TFs) (Fig.~\ref{fig:tfs}).
Rather than making the strong assumption that all the provided TFs preserve class labels, as existing approaches do, we assume a weaker form of class invariance which enables us to use \textit{unlabeled} data to learn a generative model over transformation sequences.
We then propose two representative model classes to handle modeling both commutative and non-commutative transformations.

\subsection{Augmentation as Sequence Modeling}

In our approach, we represent transformations as sequences of incremental operations.
In this setting, the user provides a set of $K$ TFs, $h_i : \mathcal{X} \mapsto \mathcal{X}$, $i\in[1,K]$.
Each TF performs an incremental transformation: for example, $h_i$ could rotate an image by five degrees, swap a word in a sentence, or move a segmented tumor mass around a background mammography image (see Fig.~\ref{fig:tfs}).
In order to accommodate a wide range of such user-defined TFs, we treat them as black-box functions which need not be deterministic nor differentiable.

This formulation gives us a tractable way to tune both the parameterization and composition of the TFs in a discretized but fine-grained manner.
Our representation can be thought of as an implicit binning strategy for tuning parameterizations -- e.g. a 15 degree rotation might be represented as three applications of a five-degree rotation TF.
It also provides a direct way to represent compositions of multiple transformation operations.
This is critical as a multitude of state-of-the-art results in the literature show the importance of using compositions of more than one transformations per image~\cite{dosovitskiy2015discriminative,ciresan80deep,graham2014fractional}, which we also confirm experimentally in Section~\ref{sec:experiments}.

\subsection{Weakening the Class-Invariance Assumption}

\begin{figure}
\centering
\includegraphics[width=0.98\linewidth]{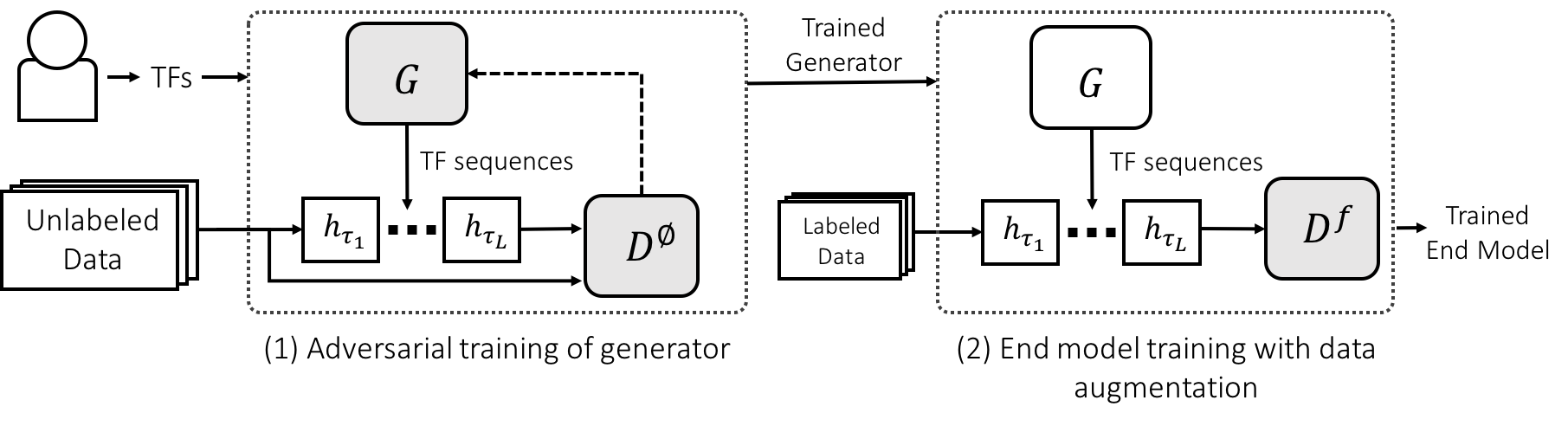}
\caption{
A high-level diagram of our method.
Users input a set of transformation functions $h_1,...,h_K$ and unlabeled data.
A generative adversarial approach is then used to train a \textit{null class} discriminator, $D^\emptyset$, and a generator, $G$, which produces TF sequences $h_{\tau_1},...,h_{\tau_L}$.
Finally, the trained generator is used to perform data augmentation for an end discriminative model $D^f$.}
\label{fig:system}
\end{figure}

Any data augmentation technique fundamentally relies on some assumption about the transformation operations' relation to the class labels.
Previous approaches make the unrealistic assumption that all provided transformation operations preserve class labels for all data points.
That is,
\begin{align}
y(h_{\tau_L} \circ\hdots\circ h_{\tau_1}(x)) = y(x) \label{assumption1}
\end{align}
for label mapping function $y$, any sequence of TF indices $\tau_1,...,\tau_L$, and \textit{all} data points $x$.
 
This assumption puts a large burden of precise specification on the user, and based on our observations, is violated by many real-world data augmentation strategies.
Instead, we consider a weaker modeling assumption.
We assume that transformation operations will not map between classes, but might destructively map data points out of the distribution of interest entirely:
\begin{align}
y(h_{\tau_L} \circ\hdots\circ h_{\tau_1}(x)) \in \{y(x), y_\emptyset\} \label{assumption2}
\end{align}
where $y_\emptyset$ represents an out-of-distribution \textit{null class}.
Intuitively, this weaker assumption is motivated by the categorical image classification setting, where we observe that transformation operations provided by the user will almost never turn, for example, a plane into a car, but may often turn a plane into an indistinguishable ``garbage'' image (Fig.~\ref{fig:inter-class}).
We are the first to consider this weaker invariance assumption, which we believe more closely matches various practical data augmentation settings of interest.
In Section~\ref{sec:experiments}, we also provide empirical evidence that this weaker assumption is useful in binary classification settings and over modalities other than image data.
Critically, it also enables us to learn a model of TF sequences using unlabeled data alone.

\subsection{Minimizing Null Class Mappings Using Unlabeled Data}

\begin{figure*}[t!]
\begin{minipage}[t]{.65\textwidth}
\centering
\include{tanda_invariance_fig}
\vspace{-8pt}
\caption{Our modeling assumption is that transformations may map out of the natural distribution of interest, but will rarely map \textit{between} classes. As a demonstration, we take images from CIFAR-10 (each row) and randomly search for a transformation sequence that best maps them to a different class (each column), according to a trained discriminative model. The matches rarely resemble the target class but often no longer look like ``normal'' images at all. Note that we consider a fixed set of user-provided TFs, not adversarially selected ones.}
\label{fig:inter-class}
\end{minipage}
\hfill
\begin{minipage}[t]{.3\textwidth}
\centering
\vspace{10pt}
\include{tanda_example_imgs_fig}
\caption{Some example transformed images generated using an augmentation generative model trained using our approach. Note that this is not meant as a comparison to Fig.~\ref{fig:inter-class}.}
\label{fig:tan-ex-imgs}
\end{minipage}
\end{figure*}

Given assumption (\ref{assumption2}), our objective is to learn a model $G_\theta$ which generates sequences of TF indices $\tau\in\{1,K\}^L$ with fixed length $L$, such that the resulting TF sequences $h_{\tau_1},...,h_{\tau_L}$ are not likely to map data points into $y_\emptyset$.
Crucially, this does not involve using the class labels of any data points, and so we can use unlabeled data.
Our goal is then to minimize the the probability of a generated sequence mapping unlabeled data points into the null class, with respect to $\theta$:
\begin{align}
J_{\emptyset} &= \mathbb{E}_{\tau\sim G_\theta}
                \mathbb{E}_{x\sim\mathcal{U}}\left[ 
                    P(y(h_{\tau_L}\circ\hdots\circ h_{\tau_1}(x)) = y_\emptyset)
            \right]
\end{align}
where $\mathcal{U}$ is some distribution of unlabeled data.

\paragraph*{Generative Adversarial Objective}
In order to approximate $P(y(h_{\tau_1}\circ\hdots\circ h_{\tau_L}(x)) = y_\emptyset)$, we jointly train the generator $G_{\theta}$ and a discriminative model $D_\phi^\emptyset$ using a generative adversarial network (GAN) objective~\cite{goodfellow2014generative}, now minimizing with respect to $\theta$ and maximizing with respect to $\phi$:
\begin{align}
\tilde{J}_{\emptyset} &= \mathbb{E}_{\tau\sim G_{\theta}}                 	\mathbb{E}_{x\sim\mathcal{U}}\left[ 
                    \log(1 - D_\phi^\emptyset(h_{\tau_L}\circ\hdots\circ h_{\tau_1}(x)))
                \right]
          + \mathbb{E}_{x'\sim\mathcal{U}}\left[ \log(D_\phi^\emptyset(x')) \right] \label{obj1}
\end{align}
As in the standard GAN setup, the training procedure can be viewed as a minimax game in which the discriminator’s goal is to assign low values to transformed, out-of-distribution data points and high values to real in-distribution data points, while simultaneously, the generator's goal is to generate transformation sequences which produce data points that are indistinguishable from real data points according to the discriminator.
For $D^\emptyset_\phi$, we use an all-convolution CNN as in~\cite{radford2015unsupervised}. For further details, see \appendixorsupp{sec:tan-details}.

\paragraph*{Diversity Objective}
An additional concern is that the model will learn a variety of null transformation sequences (e.g. rotating first left than right repeatedly).
Given the potentially large state-space of actions, and the black-box nature of the user-specified TFs, it seems infeasible to hard-code sets of inverse operations to avoid.
To mitigate this, we instead consider a second objective term:
\begin{align}
J_{d} &= \mathbb{E}_{\tau\sim G_\theta} 
             \mathbb{E}_{x\sim\mathcal{U}}\left[
                 d(h_{\tau_L}\circ\hdots\circ h_{\tau_1}(x), x)
             \right] \label{obj2}
\end{align}
where $d:\mathcal{X}\times\mathcal{X}\to\mathbb{R}$ is some distance function.
For $d$, we evaluated using both distance in the raw input space, and in the feature space learned by the final pre-softmax layer of the discriminator $D_\phi^\emptyset$.
Combining eqns.~\ref{obj1} and~\ref{obj2}, our final objective is then $J = \tilde{J}_{\emptyset} + \alpha J_{d}^{-1}$ where $\alpha> 0$ is a hyperparameter.
We minimize $J$ with respect to $\theta$ and maximize with respect to $\phi$.

\subsection{Modeling Transformation Sequences}

We now consider two model classes for $G_\theta$:

\paragraph*{Independent Model}
We first consider a \textit{mean field} model in which each sequential TF is chosen independently.
This reduces our task to one of learning $K$ parameters, which we can think of as representing the task-specific ``accuracies'' or ``frequencies'' of each TF.
For example, we might want to learn that elastic deformations or swirls should only rarely be applied to images in CIFAR-10, but that small rotations can be applied frequently.
In particular, a mean field model also provides a simple way of effectively learning stochastic,  discretized parameterizations of the TFs.
For example, if we have a TF representing five-degree rotations, \texttt{Rotate5Deg}, a marginal value of $P_{G_\theta}(\texttt{Rotate5Deg}) = 0.1$ could be thought of as roughly equivalent to learning to rotate $0.5L$ degrees on average.

\paragraph*{State-Based Model}
There are important cases, however, where the independent representation learned by the mean field model could be overly limited.
In many settings, certain TFs may have very different effects depending on which other TFs are applied with them.
As an example, certain similar pairs of image transformations might be overly lossy when applied together, such as a blur and a zoom operation, or a brighten and a saturate operation.
A mean field model could not represent such disjunctions as these.
Another scenario where an independent model fails is where the TFs are non-commutative, such as with lossy operators (e.g. image transformations which use aliasing).
In both of these cases, modeling the sequences of transformations could be important.
Therefore we consider a long short-term memory (LSTM) network as as a representative sequence model.
The output from each cell of the network is a distribution over the TFs.
The next TF in the sequence is then sampled from this distribution, and is fed as a one-hot vector to the next cell in the network.

\section{Learning a Transformation Sequence Model}
\label{sec:rl}

The core challenge that we now face in learning $G_\theta$ is that it generates sequences over TFs which are not necessarily differentiable or deterministic.
This constraint is a critical facet of our approach from the usability perspective, as it allows users to easily write TFs as black-box scripts in the language of their choosing, leveraging arbitrary subfunctions, libraries, and methods.
In order to work around this constraint, we now describe our model in the syntax of reinforcement learning (RL), which provides a convenient framework and set of approaches for handling computation graphs with non-differentiable or stochastic nodes~\cite{schulman2015gradient}.

\paragraph*{Reinforcement Learning Formulation}

Let $\tau_i$ be the index of the $i$th TF applied, and $\tilde{x}_i$ be the resulting incrementally transformed data point. Then we consider $s_t = \left({x, \tilde{x}_1, \tilde{x}_2,..., \tilde{x}_t, \tau_1, ...., \tau_t}\right)$ as the state after having applied $t$ of the incremental TFs.
Note that we include the incrementally transformed data points $\tilde{x}_1, ..., \tilde{x}_t$ in $s_t$ since the TFs may be stochastic.
Each of the model classes considered for $G_\theta$ then uses a different \textit{state representation} $\hat{s}$.
For the mean field model, the state representation used is $\hat{s}_t^{\text{MF}} = \emptyset$.
For the LSTM model, we use $\hat{s}_t^{\text{LSTM}} = \textsf{LSTM}(\tau_t, s_{t-1})$, the state update operation performed by a standard LSTM cell parameterized by $\theta$.

\paragraph*{Policy Gradient with Incremental Rewards}
Let $\ell_t(x,\tau) = \log (1 -D_\phi^\emptyset(\tilde{x}_t) )$ be the \textit{cumulative loss} for a data point $x$ at step $t$, with $\ell_0(x) = \ell_0(x,\tau) \equiv \log (1-D_\phi^\emptyset(x))$.
Let $R(s_t) = \ell_t(x,\tau) - \ell_{t-1}(x,\tau)$ be the \textit{incremental reward}, representing the difference in discriminator loss at incremental transformation step $t$.
We can now recast the first term of our objective $\tilde{J}_\emptyset$ as an expected sum of incremental rewards:
\begin{align}
U(\theta) &\equiv \mathbb{E}_{\tau\sim G_{\theta}}
                \mathbb{E}_{x\sim\mathcal{U}}\left[ 
                    \log(1 - D_\phi^\emptyset(h_{\tau_1}\circ\hdots\circ h_{\tau_L}(x)))
                \right]
             = \mathbb{E}_{\tau\sim G_{\theta}}
                \mathbb{E}_{x\sim\mathcal{U}}\left[ 
                    \ell_0(x) + \sum_{t=1}^L R(s_t)
                \right]
\end{align}
We omit $\ell_0$ in practice, equivalent to using the loss of $x$ as a baseline term.
Next, let $\pi_\theta$ be the stochastic transition policy implictly defined by $G_\theta$. We compute the recurrent policy gradient~\cite{wierstra2010recurrent} of the objective $U(\theta)$ as:
\begin{align}
\nabla_\theta U(\theta) &= \mathbb{E}_{\tau\sim G_{\theta}}
                \mathbb{E}_{x\sim\mathcal{U}}\left[ 
                    \sum_{t=1}^L R(s_t) \nabla_\theta \log \pi_\theta(\tau_t\ |\ \hat{s}_{t-1})
                \right]
\end{align}
Following standard practice, we approximate this quantity by sampling batches of $n$ data points and $m$ sampled action sequences per data point.
We also use standard techniques of discounting with factor $\gamma\in [0,1]$ and considering only future rewards~\cite{greensmith2004variance}.
See \appendixorsupp{sec:rl-details} for details.

\section{Related Work}
\label{sec:related-work}

We now review related work, both to motivate comparisons in the experiments section and to present complementary lines of work.

\paragraph*{Heuristic Data Augmentation}
Most state-of-the-art image classification pipelines use some limited form of data augmentation~\cite{graham2014fractional,dosovitskiy2015discriminative}.
This generally consists of applying crops, flips, or small affine transformations, in fixed order or at random, and with parameters drawn randomly from hand-tuned ranges.
In addition, various studies have applied heuristic data augmentation techniques to modalities such as audio~\cite{uhlich2017improving} and text~\cite{lu2006enhancing}.
As reported in the literature, the selection of these augmentation strategies can have large performance impacts, and thus can require extensive selection and tuning by hand~\cite{ciresan80deep,dosovitskiy2015discriminative} (see \appendixorsupp{sec:lit-review} as well).

\paragraph*{Interpolation-Based Techniques}
Some techniques have explored generating augmented training sets by interpolating between labeled data points.
For example, the well-known SMOTE algorithm applies this basic technique for oversampling in class-imbalanced settings~\cite{chawla2002smote}, and recent work explores using a similar interpolation approach in a learned feature space~\cite{devries2017dataset}.
\cite{hauberg2016dreaming} proposes learning a class-conditional model of diffeomorphisms interpolating between nearest-neighbor labeled data points as a way to perform augmentation.
We view these approaches as complementary but orthogonal, as our goal is to directly exploit user domain knowledge of class-invariant transformation operations.

\paragraph*{Adversarial Data Augmentation}
Several lines of recent work have explored techniques which can be viewed as forms of data augmentation that are adversarial with respect to the end classification model.
In one set of approaches, transformation operations are selected adaptively from a given set in order to maximize the loss of the end classification model being trained~\cite{teo2008convex,fawzi2016adaptive}.
These procedures make the strong assumption that all of the provided transformations will preserve class labels, or use bespoke models over restricted sets of operations~\cite{sixt2016rendergan}.
Another line of recent work has showed that augmentation via small adversarial linear perturbations can act as a regularizer~\cite{goodfellow2014explaining,miyato2015distributional}.
While complimentary, this work does not consider taking advantage of non-local transformations derived from user knowledge of task or domain invariances.

Finally, generative adversarial networks (GANs)~\cite{goodfellow2014generative} have recently made great progress in learning complete data generation models from unlabeled data.
These can be used to augment labeled training sets as well.
Class-conditional GANs~\cite{baluja2017adversarial,mirza2014conditional} generate artificial data points but require large sets of labeled training data to learn from.
Standard unsupervised GANs can be used to generate additional out-of-class data points that can then augment labeled training sets~\cite{salimans2016improved,springenberg2015unsupervised}.
We compare our proposed approach with these methods empirically in Section~\ref{sec:experiments}.

\section{Experiments}
\label{sec:experiments}
We experimentally validate the proposed framework by learning augmentation models for several benchmark and real-world data sets, exploring both image recognition and natural language understanding tasks.
Our focus is on the performance of end classification models trained on labeled datasets augmented with our approach and others used in practice.
We also examine robustness to user misspecification of TFs, and sensitivity to core hyperparameters.

\subsection{Datasets and Transformation Functions}
\label{sec:datasets}

\paragraph*{Benchmark Image Datasets}
We ran experiments on the MNIST~\cite{lecun1998gradient} and CIFAR-10~\cite{krizhevsky2009learning} datasets, using only a subset of the class labels to train the end classification models and treating the rest as unlabeled data.
We used a generic set of TFs for both MNIST and CIFAR-10: small rotations, shears, central swirls, and elastic deformations.
We also used morphologic operations for MNIST, and adjustments to hue, saturation, contrast, and brightness for CIFAR-10.

\paragraph*{Benchmark Text Dataset}
We applied our approach to the \textit{Employment} relation extraction subtask from the NIST
Automatic Content Extraction (ACE) corpus~\cite{doddington2004automatic}, where the goal is to identify mentions of employer-employee relations in news articles.
Given the standard class imbalance in information extraction tasks like this, we used data augmentation to oversample the minority positive class.
The flexibility of our TF representation allowed us to take a straightforward but novel approach to data augmentation in this setting.
We constructed a trigram language model using the ACE corpus and Reuters Corpus Volume I~\cite{lewis2004rcv1} from which we can sample a word conditioned on the preceding words.
We then used this model as the basis for a set of TFs that select words to swap based on the part-of-speech tag and location relative to entities of interest (see \appendixorsupp{sec:exp-details-text} for details).

\paragraph*{Mammography Tumor-Classification Dataset}
To demonstrate the effectiveness of our approach on real-world applications, we also considered the task of classifying benign versus malignant tumors from images in the Digital Database for Screening Mammography (DDSM) dataset~\cite{heath2000digital, Clark2013, LeeDDSM}, which is a class-balanced dataset consisting of 1506  labeled mammograms.
In collaboration with domain experts in radiology, we constructed two basic TF sets.
The first set consisted of standard image transformation operations subselected so as not to break class-invariance in the mammography setting.
For example, brightness operations were excluded for this reason.
The second set consisted of both the first set as well as several novel segmentation-based transplantation TFs.
Each of these TFs utilized the output of an unsupervised segmentation algorithm to isolate the tumor mass, perform a transformation operation such as rotation or shifting, and then stitch it into a randomly-sampled benign tissue image.
See Fig.~\ref{fig:tfs} (right panel) for an illustrative example, and \appendixorsupp{sec:exp-details-ddsm} for further details.

\subsection{End Classifier Performance} 
\label{subsec:endperf}

We evaluated our approach by using it to augment labeled training sets for the tasks mentioned above, and show that we achieve strong gains over heuristic baselines.
In particular, for a given set of TFs, we evaluate the performance of mean field (\textit{MF}) and LSTM generators trained using our approach against two standard data augmentation techniques used in practice.
The first (\textit{Basic}) consists of applying random crops to images, or performing simple minority class duplication for the ACE relation extraction task.
The second (\textit{Heur.}) is the standard heuristic approach of applying random compositions of the given set of transformation operations, the most common technique used in practice~\cite{ciresan80deep,graham2014fractional,he2016deep}.
For both our approaches (\textit{MF} and \textit{LSTM}) and \textit{Heur.}, we additionally use the same random cropping technique as in the \textit{Basic} approach.
We present these results in Table~\ref{tbl:main-results}, where we report test set accuracy (or F1 score for ACE), and use a random subsample of the available labeled training data.
Additionally, we include an extra row for the DDSM task highlighting the impact of adding domain-specific (\textit{DS}) TFs -- the segmentation-based operations described above -- on performance.

In Table~\ref{tbl:gan-comparisons} we additionally compare to two related generative-adversarial methods, the Categorical GAN (CatGAN)~\cite{springenberg2015unsupervised}, and the semi-supervised GAN (SS-GAN) from~\cite{salimans2016improved}.
Both of these methods use GAN-based architectures trained on unlabeled data to generate new out-of-class data points with which to augment a labeled training set.
Following their protocol for CIFAR-10, we train our generator on the full set of unlabeled data, and our end discriminator on ten disjoint random folds of the labeled training set not including the validation set (i.e. $n=4000$ each), averaging the results.

\begin{table*}[htbp!]
\setlength\tabcolsep{5pt}
\parbox{.67\linewidth}{
\centering
\begin{tabular}{lc|ccccc}
\toprule
{\bf Task} & {\bf \%} &{\bf None} & {\bf Basic} & {\bf Heur.} & {\bf MF} & {\bf LSTM}\\
\midrule
MNIST & 1 & 90.2 & 95.3 & 95.9 & 96.5 & \bf{96.7} \\
 & 10 & 97.3 & 98.7 & 99.0 & \textbf{99.2} & 99.1 \\ \midrule
CIFAR-10 & 10 & 66.0 & 73.1 & 77.5 & 79.8 & \bf{81.5} \\
 & 100 & 87.8 & 91.9 & 92.3 & \bf{94.4} & 94.0 \\\midrule
ACE (F1) & 100 & 62.7 & 59.9 & 62.8 & 62.9 & \bf{64.2} \\ \midrule
DDSM & \multirow{2}{*}{10} & \multirow{2}{*}{57.6}  & \multirow{2}{*}{58.8} & 59.3 & 58.2 & 61.0 \\
DDSM + DS & &  &  & 53.7 & 59.9 & \bf{62.7} \\
\bottomrule 
\end{tabular}
\caption{Test set performance of end models trained on subsamples of the labeled training data (\textit{\%}), not including validation splits, using various data augmentation approaches. \textit{None} indicates performance with no augmentation. All tasks are measured in accuracy, except ACE which is measured by F1 score.}
\label{tbl:main-results}
}
\hfill
\parbox{.28\linewidth}{
\centering
\begin{tabular}{cc}
\toprule
{\bf Model} & {\bf Acc. (\%)} \\
\midrule
CatGAN & $80.42 \pm 0.58$\\
SS-GAN & $81.37 \pm 2.32$\\
\textbf{LSTM} & \textbf{$81.47 \pm 0.46$}\\
\bottomrule 
\end{tabular}
\caption{Reported end model accuracies, averaged across 10\% subsample folds, on CIFAR-10 for comparable GAN methods.}
\label{tbl:gan-comparisons}
}
\end{table*}

\begin{figure}
\centering
\begin{subfigure}{.48\textwidth}
\includegraphics[width=\linewidth]{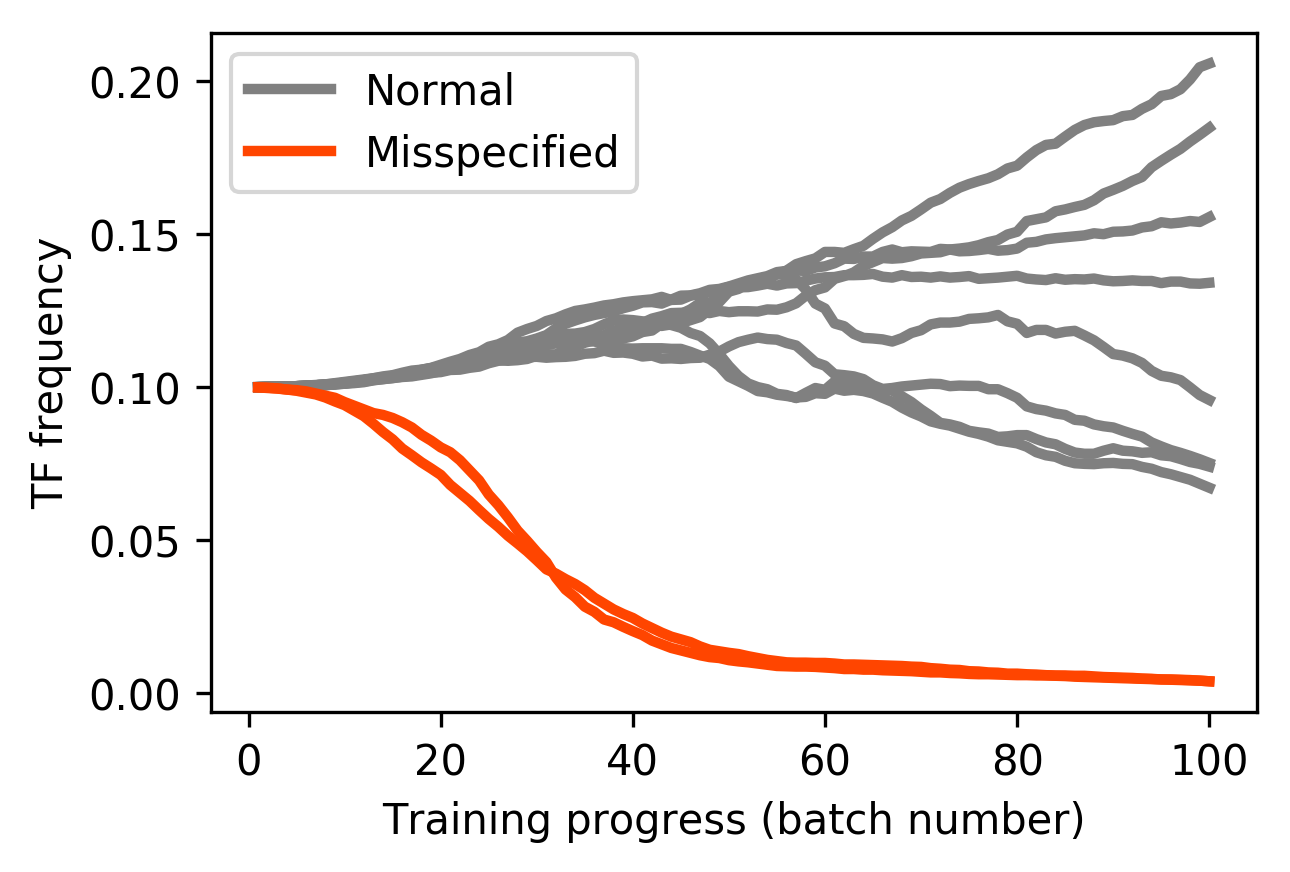}
\caption{}
\label{fig:robustness}
\end{subfigure}%
\hfill
\begin{subfigure}{.48\textwidth}
\includegraphics[width=\linewidth]{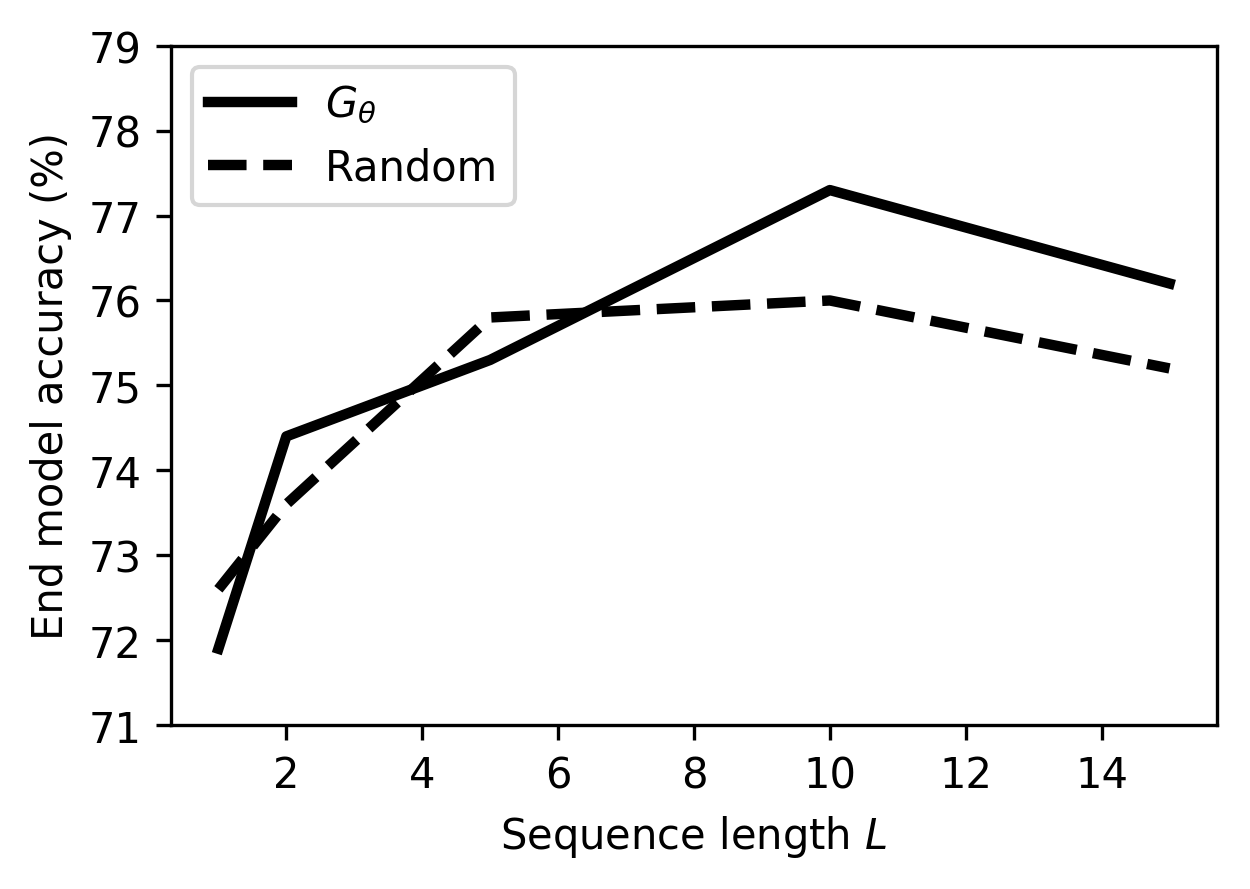}
\caption{}
\label{fig:seq-len}
\end{subfigure}%
\caption{
(a) Learned TF frequency parameters for misspecified and normal TFs on MNIST. The mean field model correctly learns to avoid the misspecified TFs.
(b) Larger sequence lengths lead to higher end model accuracy on CIFAR-10, while random performs best with shorter sequences, according to a sequence length calibration experiment.
}
\label{fig:plots}
\end{figure}

In all settings, we train our TF sequence generator on the full set of unlabeled data.
We select a fixed sequence length for each task via an initial calibration experiment (Fig.~\ref{fig:seq-len}).
We use $L=5$ for ACE, $L=7$ for DDSM + DS, and $L=10$ for all other tasks.
We note that our findings here mirrored those in the literature, namely that compositions of multiple TFs lead to higher end model accuracies.
We selected hyperparameters of the generator via performance on a validation set.
We then used the trained generator to transform the entire training set at each epoch of end classification model training.
For MNIST and DDSM we use a four-layer all-convolutional CNN, for CIFAR10 we use a 56-layer ResNet~\cite{he2016deep}, and for ACE we use a bi-directional LSTM.
Additionally, we incorporate a basic transformation regularization term as in~\cite{SajjadiJT16a} (see \appendixorsupp{sec:exp-details-end-model}), and train for the last ten epochs without applying any transformations as in~\cite{graham2014fractional}.
In all cases, we use hyperparameters as reported in the literature.
For further details of generator and end model training see the \appendixorsupp{sec:tan-details}.

We see that across the applications studied, our approach outperforms the heuristic data augmentation approach most commonly used in practice.
Furthermore, the LSTM generator outperforms the simple mean field one in most settings, indicating the value of modeling sequential structure in data augmentation.
In particular, we realize significant gains over standard heuristic data augmentation on CIFAR-10, where we are competitive with comparable semi-supervised GAN approaches, but with significantly smaller variance.
We also train the same CIFAR-10 end model using the full labeled training dataset, and again see strong relative gains (2.1 pts. in accuracy over heuristic), coming within 2.1 points of the current state-of-the-art~\cite{huang2016densely} using our much simpler end model.

On the ACE and DDSM tasks, we also achieve strong performance gains, showing the ability of our method to productively incorporate more complex transformation operations from domain expert users.
In particular, in DDSM we observe that the addition of the segmentation-based TFs causes the heuristic augmentation approach to perform significantly worse, due to a large number of new failure modes resulting from combinations of the segmentation-based TFs -- which use gradient-based blending -- and the standard TFs such as zoom and rotate.
In contrast, our LSTM model learns to avoid these destructive subsequences and achieves the highest score, resulting in a 9.0 point boost over the comparable heuristic approach.

\paragraph*{Robustness to TF Misspecification}
\label{subsec:robustness-tfs}

One of the high-level goals of our approach is to enable an easier interface for users by not requiring that the TFs they specify be completely class-preserving.
The lack of any assumption of well-specified transformation operations in our approach, and the strong empirical performance realized, is evidence of this robustness.
To additionally illustrate the robustness of our approach to misspecified TFs, we train a mean field generator on MNIST using the standard TF set, but with two TFs (shear operations) parameterized so as to map almost all images to the null class.
We see in Fig.~\ref{fig:robustness} that the generator learns to avoid applying the misspecified TFs (red lines) almost entirely.

\section{Conclusion and Future Work}
\label{sec:conclusion}

We presented a method for learning how to parameterize and compose user-provided black-box transformation operations used for data augmentation.
Our approach is able to model arbitrary TFs, allowing practitioners to leverage domain knowledge in a flexible and simple manner.
By training a generative sequence model over the specified transformation functions using reinforcement learning in a GAN-like framework, we are able to generate realistic transformed data points which are useful for data augmentation.
We demonstrated that our method yields strong gains over standard heuristic approaches to data augmentation for a range of applications, modalities, and complex domain-specific transformation functions.
There are many possible future directions of research for learning data augmentation strategies in the proposed model, such as conditioning the generator's stochastic policy on a featurized version of the data point being transformed, and generating TF sequences of dynamic length.
More broadly, we are excited about further formalizing  data augmentation as a novel form of weak supervision, allowing users to directly encode domain knowledge about invariants into machine learning models.

\paragraph*{Acknowledgements}
We would like to thank Daniel Selsam, Ioannis Mitliagkas, Christopher De Sa, William Hamilton, and Daniel Rubin for valuable feedback and conversations.
We gratefully acknowledge the support of
the Defense Advanced Research Projects Agency (DARPA) SIMPLEX program under No. N66001-15-C-4043,
the DARPA D3M program under No. FA8750-17-2-0095,
DARPA programs No. FA8750-12-2-0335 and FA8750-13-2-0039,
DOE 108845,
National Institute of Health (NIH) U54EB020405,
the Office of Naval Research (ONR) under awards No. N000141210041 and No. N000141310129,
the Moore Foundation,
the Okawa Research Grant,
American Family Insurance,
Accenture,
Toshiba,
and Intel.
This research was also supported in part by affiliate members and other supporters of the Stanford DAWN project: Intel, Microsoft, Teradata, and VMware.
This material is based on research sponsored by DARPA under agreement number FA8750-17-2-0095.
The U.S. Government is authorized to reproduce and distribute reprints for Governmental purposes notwithstanding any copyright notation thereon. 
Any opinions, findings, and conclusions or recommendations expressed in this material are those of the authors and do not necessarily reflect the views, policies, or endorsements, either expressed or implied, of DARPA, AFRL, NSF, NIH, ONR, or the U.S. Government.

\bibliography{tanda}{}
\bibliographystyle{abbrv}

\appendix

\section{Additional Background}
\subsection{Review of Data Augmentation Use in State-of-the-Art}
\label{sec:lit-review}

To underscore both the omnipresence and diversity of heuristic data augmentation in practice, we compiled a list of the top ten models for the well documented CIFAR-10 and CIFAR-100 tasks (Table~\ref{tbl:lit-review}). We see that in 10 out of 10 of the top CIFAR-10 results and 9 out of 10 of the top CIFAR-100 results use data augmentation, for average boosts (when reported) of 3.71 and 13.39 points in accuracy, respectively. Moreover, we see that while some sets of papers inherit a simple data augmentation strategy from prior work (in particular, all the recent ResNet variants), there are still a large variety of approaches. And in general, the particular choice of data augmentation strategy is widely reported to have large effects on performance.

Note that the below table is compiled from a well-known online compendium~\footnote{\url{}} and from the latest CVPR best paper~\cite{huang2016densely} (indicated by a *) which achieves new state-of-the-art results. We compile it for illustrative purposes and it is not necessarily comprehensive. Also note that we selected CIFAR-10/100 both as a representative and well-studied task, but also due to the availability of published results. For competitions such as ImageNet, although data augmentation is widely reported to be critical, many top results are reported very opaquely, with little described about implementation details such as data augmentation.

\begin{table*}[htbp!]
\setlength\tabcolsep{5pt}
\centering
\begin{tabular}{l c p{4cm} c c p{4cm}}
\toprule
{\bf Dataset} & {\bf Pos.} &{\bf Name} & {\bf \small{Err. w/DA}} & {\bf \small{Err. w/o DA}} & {\bf Notes}\\
\midrule
\multirow{10}{*}{\small{CIFAR-10}} & 1 & DenseNet & 3.46 & - & Random shifts, flips \\
	 & 2 & Fractional Max-Pooling & 3.47 & - & \small{Randomized mix of translations, rotations, reflections, stretching, shearing, and random RGB color shift operations}\\
	 \cmidrule{2-6}
	 & 3* & Wide ResNet & 4.17 & - & Random shifts, flips\\
	 \cmidrule{2-6}
	 & 4 & Striving for Simplicity: The All Convolutional Net & 4.41 & 9.08 & “Heavy” augmentation: images expanded, then scaled, rotated, color shifted randomly\\
	 \cmidrule{2-6}
	 & 5* & FractalNet & 4.60 & 7.33 & Random shifts, flips\\
	 \cmidrule{2-6}
	 & 6* & ResNet (1001-Layer) & 4.62 & 10.56 & Random shifts, flips\\
	 \cmidrule{2-6}
	 & 7* & ResNet with Stochastic Depth (1202-Layer) & 4.91 & - & Random shifts, flips\\
	 \cmidrule{2-6}
	 & 8 & All You Need is a Good Init & 5.84 & - & Random shifts, flips\\
	 \cmidrule{2-6}
	 & 9 & Generalizing Pooling Functions in Convolutional Neural Networks: Mixed, Gated, and Tree & 6.05 & 7.62 & Flips, random shifts, other simple ones\\
	 \cmidrule{2-6}
	 & 10 & Spatially-Sparse Convolutional Neural Networks & 6.28 & - & Affine transformations\\
	 \toprule
\multirow{10}{*}{\small{CIFAR-100}} & 1* & DenseNet & 17.18 & - & Random shifts, flips\\
     \cmidrule{2-6}
	 & 2* & Wide ResNets & 20.50 & - & Random shifts, flips\\
	 \cmidrule{2-6}
	 & 3* & ResNet (1001-Layer) & 22.71 & 33.47 & Random shifts, flips\\
	 \cmidrule{2-6}
	 & 4* & FractalNet & 23.30 & 35.34 & Random shifts, flips\\
	 \cmidrule{2-6}
	 & 5 & Fast and Accurate Deep Network Learning by Exponential Linear Units & - & 24.28 & \\
	 \cmidrule{2-6}
	 & 6 & Spatially-Sparse Convolutional Neural Networks & 24.3 & - & Affine transformations\\
	 \cmidrule{2-6}
	 & 7* & ResNet with Stochastic Depth (1202-Layer) & 24.58 & 37.80 & Random shifts, flips\\
	 \cmidrule{2-6}
	 & 8 & Fractional Max-Pooling & 26.39 & - & Randomized mix of translations, rotations, reflections, stretching, and shearing operations, and random RGB color shifts\\
	 \cmidrule{2-6}
	 & 9* & ResNet (110-Layer) & 27.22 & 44.74 & Random shifts, flips\\
	 \cmidrule{2-6}
	 & 10 & Scalable Bayesian Optimization Using Deep Neural Networks & 27.4 & - & Hue, saturation, scalings, horizontal flips\\
\bottomrule 
\end{tabular}
\caption{Current state-of-the-art image classification models as ranked by reported performance on the CIFAR-10 and CIFAR-100 tasks, and their error with (\textit{Err. w/ DA}) and without (\textit{Err. w/o DA}) data augmentation. We include both scores and particular data augmentation techniques when reported, although the latter is rarely reported with great precision.}
\label{tbl:lit-review}
\end{table*}

\section{Additional Experiments}
\subsection{Synthetic Data Examples}

\paragraph*{Simple Synthetic Setup}
As both a diagnostic tool and as a simple way to probe the properties of our approach, we construct a simple synthetic dataset consisting of points in two dimensions, uniformly selected in a ball of radius $r=1$ around the origin.
We then consider various displacement vectors as our TFs.
We consider the same two generative models as in our main experiments -- mean field and LSTM -- and use either a basic fully-connected two-layer neural network or an oracle discriminator $f(x) = 1\{||x|| < 1\}$.

\begin{figure}[H]
\centering
\begin{subfigure}{.33\textwidth}
  \centering
  \includegraphics[width=1.0\linewidth]{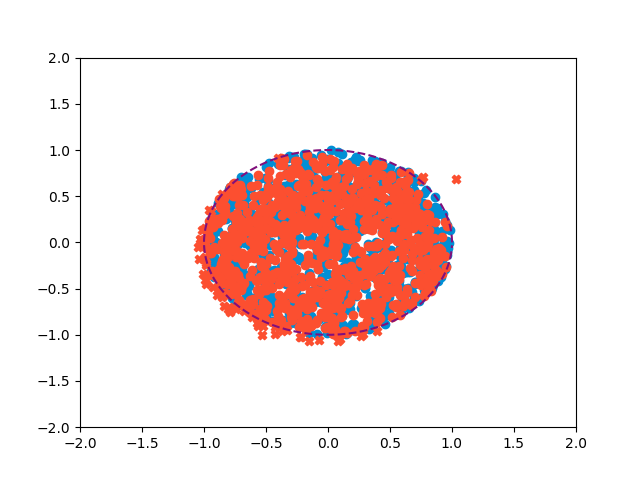}
  \caption{Mean field model on TF set 1}
  \label{fig:sub1}
\end{subfigure}%
\begin{subfigure}{.33\textwidth}
  \centering
  \includegraphics[width=1.0\linewidth]{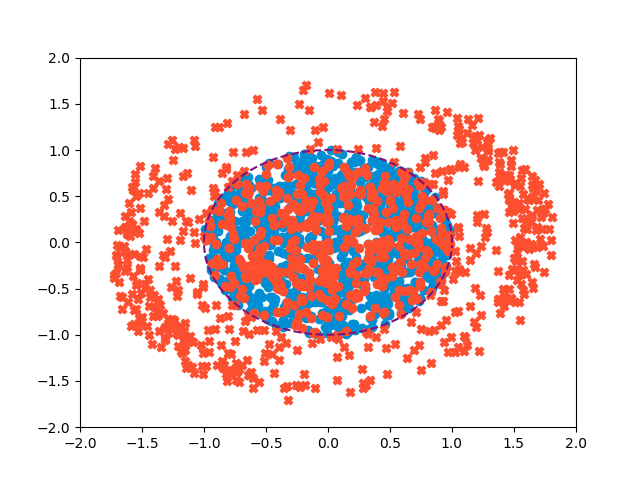}
  \caption{Mean field model on TF set 2}
  \label{fig:sub2}
\end{subfigure}
\begin{subfigure}{.33\textwidth}
  \centering
  \includegraphics[width=1.0\linewidth]{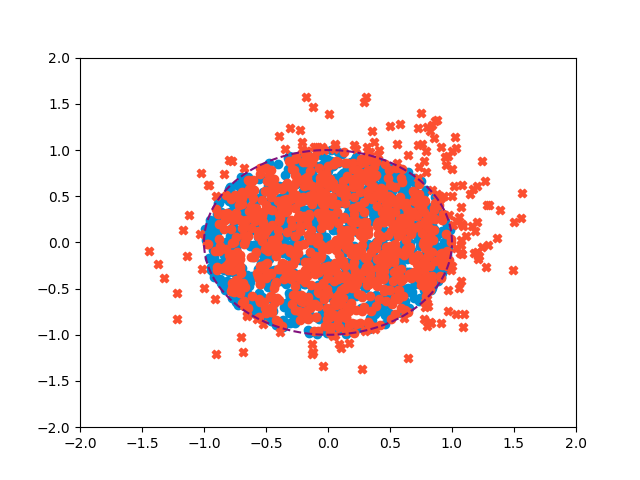}
  \caption{LSTM model on TF set 2}
  \label{fig:sub3}
\end{subfigure}
\caption{Original data points (blue) are transformed using sequences of vector displacement TFs ($L=10$) drawn from $G_\theta$, producing augmented data points (red). $G_\theta$ is either a mean field model or an LSTM, trained with an orcale discriminator $D^\emptyset$ for 15 epochs.}
\label{synthetics}
\end{figure}

\paragraph*{Synthetic Experiments}
In this setting, we define $y_\emptyset (x) = 1\{ ||x|| \ge 1 \}$, and consider two different TF sets:
\begin{compactenum}
\item \textit{Good vs. Bad TFs:}
In a first toy scenario we consider TFs which are vector displacements of random direction, with magnitude drawn from one of two distributions, $\mathcal{N}(\mu_1, \sigma_1)$ or $\mathcal{N}(\mu_2, \sigma_2)$, where $\mu_1 > 1 > \mu_2$.
In other words, the model should learn not to select certain individual TFs.

\item \textit{Lossy TFs:}
We consider a second toy setting where random-direction displacement TFs have magnitude drawn uniformly from $\mathcal{N}(\mu, \sigma)$, $\mu < 1$; however the magnitude of each TF decays exponentially with the distance a point is outside of the unit ball.
This simulates the setting where TFs are irrecoverably lossy when applied in certain sequences.
\end{compactenum}
As expected, we see that while the mean field model is able to model the first setting (Figure~\ref{fig:sub1}), it fails to adequately represent the second one (Figure~\ref{fig:sub2}), whereas the RNN model is able to (Figure~\ref{fig:sub3}).

\subsection{Robustness to Transformed Test Data}
\begin{figure}[H]
\centering
\includegraphics[width=0.8\linewidth]{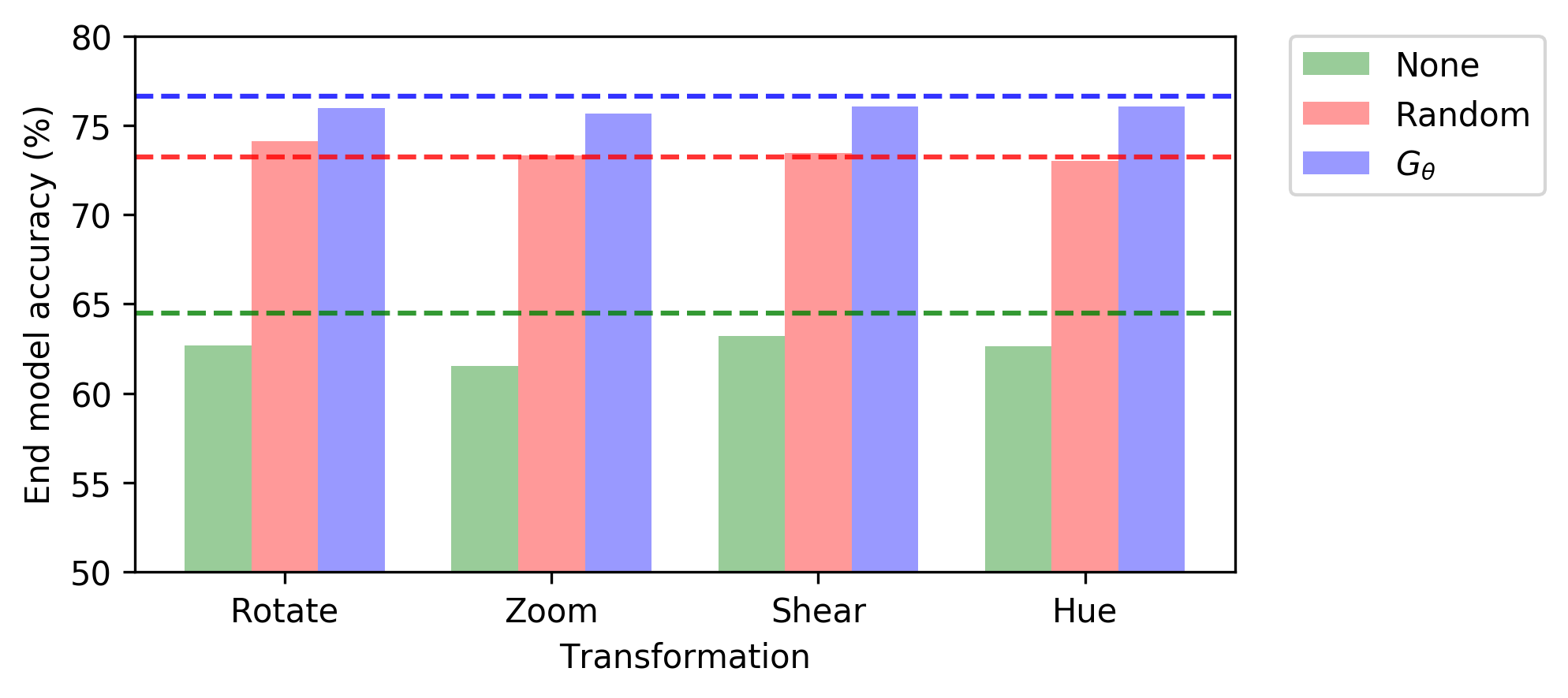}
\caption{Accuracy scores on random 10\% subsamples of test data (dotted lines) and on versions augmented with a single transformation (vertical bars) with parameters drawn uniformly at random.}
\label{fig:end_robust}
\end{figure}

We run a simple experiment to test the robustness of the trained end classification models to the individual TFs in the TF sets used.
Specifically, on CIFAR-10 we create ten transformed copies a 10\% subsample of the test data by transforming with a single TF, and then test the end model on this set.
We compare our approach with heuristic random  augmentation and no data augmentation of the model during training, and consider rotations, zooms, shears, and hue shifts.
Results are presented in Figure~\ref{fig:end_robust}.

We can consider evaluating the results in terms of absolute robustness -- i.e. model accuracy -- and relative robustness, i.e. the change in model score when applied to the transformed test set.
Roughly we see that our approach is most absolutely robust.
Random appears to be most relatively robust, in particular on larger transformations, which we hypothesize our approach mostly learned to avoid applying during training.

\section{Reinforcement Learning Formulation Details}
\label{sec:rl-details}
\subsection{Variance reduction methods}
In Section~\ref{sec:rl}, the vanilla policy gradient of our objective was given as
\begin{align*}
\nabla_\theta U(\theta) &= \mathbb{E}_{\tau\sim G_{\theta}}
                \mathbb{E}_{x\sim\mathcal{U}}\left[ 
                    \sum_{t=1}^L R(s_t) \nabla_\theta \log \pi_\theta(\tau_t\ |\ \hat{s}_{t-1})
                \right]
\end{align*}
Noting that actions $\tau_t$ only impact future outcomes, following standard practice, we apply only future rewards in order to reduce variance:
\begin{align*}
\nabla_\theta U(\theta) &= \mathbb{E}_{\tau\sim G_{\theta}}
                \mathbb{E}_{x\sim\mathcal{U}}\left[ 
                    \sum_{t=1}^L \nabla_\theta \log \pi_\theta(\tau_t\ |\ \hat{s}_{t-1}) \sum_{t'=t}^L \gamma^{t'-t} R(s_{t'}) 
                \right]
\end{align*}
where $\gamma\in[0,1]$ is a discounting factor.
Additionally, we use a baseline term $b_t$ to reduce variance when estimating $\nabla_\theta U(\theta)$:
\begin{align*}
\nabla_\theta U(\theta) &= \mathbb{E}_{\tau\sim G_{\theta}}
                \mathbb{E}_{x\sim\mathcal{U}}\left[ 
                    \sum_{t=1}^L \nabla_\theta \log \pi_\theta(\tau_t\ |\ \hat{s}_{t-1}) \left(\left(\sum_{t'=t}^L \gamma^{t'-t} R(\hat{s}_{t'}) 
                \right) - b_t\right)\right]
\end{align*}

\subsection{Policy gradient estimation}

Using a batch of $n$ data points and $m$ sampled action sequences per data point -- given by the state representations $\{\hat{s}^{(i,j)}\}$ and action sequences $\{\tau^{(i,j)}\}$ -- the gradient estimate is computed as:
\begin{align*}
\nabla_\theta \hat{U}(\theta) &= \frac{1}{nm}\sum_{i=1}^n\sum_{j=1}^m \left[ 
                    \sum_{t=1}^L \nabla_\theta \log \pi_\theta(\tau^{(i,j)}_t\ |\ \hat{s}^{(i,j)}_{t-1}) \left(\left(\sum_{t'=t}^L \gamma^{t'-t} R(\hat{s}^{(i,j)}_{t'}) 
                \right) - \hat{b}_t\right)\right]
\end{align*}
where the baseline term $\hat{b}_t$ is also computed using the batch:
\begin{align*}
\hat{b}_t = \frac{1}{nm}\sum_{i=1}^n\sum_{j=1}^m \sum_{t'=t}^L \gamma^{t'-t} R\left(\hat{s}^{(i,j)}_{t'}\right)
\end{align*}

In our experiments, we fixed $n=32$ and $m=5$.

\section{Experimental Details}
\label{sec:exp-details}
\subsection{Benchmark Image Datasets}
\label{sec:exp-details-benchmark}
We use the MNIST dataset with 5000 training data points used as a validation set.  We use the following TFs:
\begin{itemize}
    \item Rotation (2.5, -2.5, 5, -5, 10, and -10 degrees)
    \item Zoom (0.9x, 1.1x)
    \item Shear (0.1, -0.1, 0.2, -0.2, 0.4, and -0.4 degrees)
    \item Swirl (0.1, -0.1, 0.2, -0.2, 0.4, and -0.4 degrees)
    \item Random elastic deformations ($\alpha$ = 1.0, 1.25, and 1.5)
    \item Erosion
    \item Dilation
\end{itemize}
For the CIFAR-10 dataset, we use the following TFs:
\begin{itemize}
    \item Rotation (2.5, -2.5, 5, -5 degrees)
    \item Zoom (0.9x, 1.1x, 0.75x, 1.25x)
    \item Shear (0.1, -0.1, 0.25, and -0.25 degrees)
    \item Swirl (0.1, -0.1, 0.25, -0.25 degrees)
    \item Hue Shift (by 0.1, -0.1, 0.25, and -0.25)
    \item Enhance contrast (by 0.75, 1.25, 0.5, and 1.5)
    \item Enhance brightness (by 0.75, 1.25, 0.5, and 1.5)
    \item Enhance color (by 0.75, 1.25, 0.5, and 1.5)
    \item Horizontal flip
\end{itemize}

For both datasets, we also applied random padding (by 4 pixels on each side) followed by random crops back to the original dimensions during training.
We note that the choice of certain TFs to use in certain datasets was deliberate -- for example, we would not expect horizontal flips or hue shifts to be appropriate in MNIST, or erosion and dilation to be useful in CIFAR-10.
However, the particular choice of parameterizations was mainly due to disjoint implementations of the two experiments.  
For further details of the TF implementations used, see our code, which will be open-sourced after the review process.

\subsection{Benchmark Text Dataset}
\label{sec:exp-details-text}
The ACE corpus consists of news articles and broadcast transcripts, all of which are pretagged with entity mentions.
The objective of the Employment relation subtask is to extract \textit{Person}-\textit{Organization} entity pairs which are implied to have an affiliation in the text.
We pose this as a binary classification problem by first identifying relation \textit{candidates}: any pair of \textit{Person}-\textit{Organization} entities which occur in the same sentence.
As noted in Section \ref{sec:experiments}, there are far more true negative candidates than true positive candidates.
The end model is trained to classify relation candidates as either true or false relations based on the raw text of the sentence in which they occur.

The language model described in Section~\ref{sec:experiments} was constructed by recording counts of unigrams following each unique trigram, bigram, and unigram in the corpus. 
Laplace smoothing was applied to the counts, and basic filtering was applied to the $n$-grams.
The sampler falls back to using bigrams then unigrams if the trigram preceding the word we want to swap was filtered out of the corpus.
We used the following TFs in all experiments:
\begin{itemize}
\item Replace a noun to the left of both entities
\item Replace a noun between the two entities
\item Replace a noun to the right of both entities
\item Replace a verb to the left of both entities
\item Replace a verb between the two entities
\item Replace a verb to the right of both entities
\item Replace an adjective to the left of both entities
\item Replace an adjective between the two entities
\item Replace an adjective to the right of both entities 
\end{itemize}

\subsection{DDSM Mammography Task}
\label{sec:exp-details-ddsm}
We use the following transformation functions:

\begin{enumerate}
\item 
Rotate Image: Rotate the entire mammogram by a deterministic angle $\theta$.  Tumor geometry is fundamentally invariant to 2-D orientation. TF run with  $\theta \in [-5^o,-2.5^o,2.5^o,5^o]$.
\item 
Zoom Image: Zoom in on the entire mammogram by a deterministic factor $\gamma$.  Tumor classification is insensitive to $\gamma$ for $\gamma$ close to one.  TF run with $\gamma \in \{0.98,1.02\}$
\item 
Enhance Image Contrast: Enhance contrast values of grayscale image by a deterministic factor $\gamma$.  Tumor classification is insensitive to $\gamma$ for $\gamma$ close to one.  TF run with $\gamma \in \{0.95,1.05\}$ 
\item
Translate and Transplant Image: Extract pixels within mass segmentation.  Perform translation of a bounding box of side length $N$ pixels about mass center by a deterministic vector $\hat{g}$.  Transplant the translated bounding box containing the mass onto a randomly sampled normal tissue image using Poisson blending.  Retains information about the mass itself within the context of a different set of normal tissue background.  The bounding box also retains information about the tissue in the tumor near field.  TF run with $N = 10$, $\hat{g}\in \{(-3,0),(3,0),(0,-3),(0,3),(0,0)\}  $.
\item 
Rotate and Transplant Image: Extract pixels within mass segmentation.  Perform rotation of a bounding box of side length $N$ pixels about mass center by a deterministic angle $\theta$.  Transplant the rotated bounding box containing the mass onto a randomly sampled normal tissue image using Poisson blending~\cite{perez2003poisson}.  Retains information about the mass itself within the context of a different set of normal tissue background.  The bounding box also retains information about the tissue in the tumor near field.  TF run with $N = 10$, $\theta \in \{-5^o,-2.5^o,2.5^o,5^o\}$.
\end{enumerate}

Note that for the Poisson blending TFs, it is important that the translation and rotation domains be specified such that excessive proximity to the boundary of the destination image does not introduce spurious gradient information into the blended image.

\subsection{Details of Generative Adversarial Network Models}
\label{sec:tan-details}

All models, both for the generator training as described in this section, and the end classification model training described next, were implemented in Tensorflow~\footnote{https://www.tensorflow.org}.

\paragraph*{Discriminator}
For image tasks, the discriminator used in the training of the generator in our approach was the same model as in~\cite{radford2015unsupervised}, an all-convolutional CNN with four convolutional layers and leaky ReLU activations.
For the text task, we used a unidirectional RNN with basic LSTM cells.

\paragraph*{Mean Field Model} The mean field model is represented simply as a length $K$ vector of unbounded variables, where $K$ is the number of TFs.
Applying the softmax function to this vector yields the TF sampling distribution.

\paragraph*{LSTM} In the LSTM generative model, we create a length-$L$ RNN with basic LSTM cells.
The input and output size for each cell is $K$.
We feed an indicator vector of the last TF used as the input to each cell, except for the first cell, which recieves a randomly initialized variable vector as its input.
The output of each cell is shifted and scaled to range from $-r$ to $r$, where $r$ is a hyperparameter.
Applying the softmax function to the shifted and scaled output yields the stochastic policy: a sampling distribution over the $K$ TFs.
In our experiments, we fix $r=2$ to avoid overfitting.

\paragraph*{Training and Model Selection Procedure}
We trained the TF sequence generators jointly with the discriminator using SGD with momentum (fixed at 0.9), in an adversarial manner as described in~\cite{goodfellow2014generative}.
We performed an initial search over the TF sequence length $L$ as described in Section~\ref{sec:experiments}, and then held it fixed at $L=10$ for all subsequent experiments.
We searched over a range of values for learning rates for the generator and discriminator, as well as for hyperparameters specific to our formulation, such as the diversity objective term coefficient $\alpha$, the diversity objective term distance metric $d$ (choosing between distance in the raw input space or in the feature space learned by the final pre-softmax layer of the discriminator), and whether or not to split the data used for the discriminator and generator training steps.

We selected final generators to use for test set evaluation by using them to augment training data for end classification models then evaluated on the validation set.
In addition, we filtered some generators out based on their loss (according to the discriminator $D_\phi^\emptyset$) as compared to that of random TF sequences.

\paragraph*{Diversity Objective}
For the diversity objective term, we tried both distance in the raw pixel-level input space and distance in the feature space learned by the final pre-softmax layer of the discriminator as choices for distance metric $d$. During training of the generators, we measured the average pairwise generalized Jaccard distance. For CIFAR-10, as an example, the final batches had an average distance of 0.52 compared to 0.86 for randomly generated sequences, which implied diversity in the learned sequences. We also computed the ratio of unique TF n-grams to total possible n-grams, and measured 0.37 compared to 0.98 for random sequences as expected.

\subsection{End Classification Models}

\paragraph*{MNIST and DDSM}
For MNIST and DDSM we use a similar architecture to the discriminator in the previous section, adapted for the multinomial classification setting: a four-layer all-convolution CNN with leaky ReLUs and batch norm.

\paragraph*{CIFAR-10}
Given the flexibility of end classifier choice with our approach, for CIFAR-10 we used a more computationally expensive but standard model: a 56-layer ResNet as described in~\cite{he2016deep}.
We used batch norm, regularization, learning rate schedule, and all other hyperparameters as reported in~\cite{he2016deep}.

\paragraph*{ACE}
The end model used for the ACE task was a bidirectional recurrent neural network using LSTM cells with attention mechanisms.
The maximum sentence length and attention window length were both 50.
Word embeddings were initialized from pretrained vectors via~\cite{bojanowski2016enriching}, and updated during training.
Hyperparameters were selected via a cursory grid search, and fixed for experiments.

\subsection{End Model Training}
\label{sec:exp-details-end-model}

\paragraph*{Basic Training Procedure with Data Augmentation}
We trained all end models using minibatch stochastic gradient descent with momentum (fixed at 0.9), using a fixed learning rate schedule set once for each model and then fixed for all experiments.
To perform data augmentation, during end classifier training we transformed some portion of each minibatch, $p_{\text{transform}}$.
For all experiments we used $p_{\text{transform}}=1.0$.
Additionally, for the last ten epochs of training, we switched to $p_{\text{transform}}=0.0$ following reported practice in the literature~\cite{graham2014fractional}.
For all other hyperparameters we used default values as reported in the respective literature, held fixed at these values for all experiments.

\paragraph*{Transformation Regularization Term}
We additionally apply a \textit{transformation regularization (TR)} term to the transformed data points for all image experiments by adding a term to the loss function which is the distance between the pre-softmax layer logits for each data point and its transformed copy, similar to the term in~\cite{SajjadiJT16a}.
Given the fact that we are producing these transformed data points anyway, incorporating this term introduces little additional overhead.

\begin{table*}[htbp!]
\centering
\begin{tabular}{lc|lcc}
\toprule
{\bf Task} & {\bf \%} &{\bf Augmentation Model} & {\bf TR Term Coefficient} & {\bf Accuracy (Dev.)}\\
\midrule
\multirow{4}{*}{CIFAR-10} & \multirow{4}{*}{10} & Heuristic & 0 & 77.4\\
 & & Heuristic & 0.1 & 77.5\\
 & & LSTM & 0 & 80.4\\
 & & LSTM & 0.1 & 81.6\\
\bottomrule 
\end{tabular}
\caption{A simple study of the effect of adding a transformation regularization (TR) term to the objective function, evaluated on a labeled validation set. We see that adding the term improves performance for both heuristic (random) TF sequences and for TF sequences generated by the trained LSTM model, and that there is a larger positive effect for the latter.}
\label{tbl:tr-ablation-study}
\end{table*}

In an early calibration experiment (Table~\ref{tbl:tr-ablation-study}), we found that introducing this regularization term (using a coefficient of $0.1$ and unlabeled data batch size of $20\%$ that of the labeled data batch size) yielded improvements in performance to the end model with both learned transformation sequences and random sequences. However, we see that the positive effect is much larger for the trained LSTM sequences (1.2 points versus 0.1 points in accuracy). We chose to subsequently keep this term fixed, viewing further calibration and exploration of this term as largely orthogonal to our central experimental questions. However, we believe that this is an extremely interesting and empirically proimising area for future study, especially given the indication that this term may be more effective when used in conjunction with a trained augmentation model such as ours.

\end{document}